# Adaptive texture energy measure method

## Ömer Faruk Ertuğrul

Electrical and Electronics Engineering, Batman University, Batman, Turkey

**Email address:**
omerfaruk.ertugrul@batman.edu.tr

**Abstract:** Recent developments in image quality, data storage, and computational capacity have heightened the need for texture analysis in image process. To date various methods have been developed and introduced for assessing textures in images. One of the most popular texture analysis methods is the Texture Energy Measure (TEM) and it has been used for detecting edges, levels, waves, spots and ripples by employing predefined TEM masks to images. Despite several successful studies, TEM has a number of serious weaknesses in use. The major drawback is; the masks are predefined therefore they cannot be adapted to image. A new method, Adaptive Texture Energy Measure Method (aTEM), was offered to overcome this disadvantage of TEM by using adaptive masks by adjusting the contrast, sharpening and orientation angle of the mask. To assess the applicability of aTEM, it is compared with TEM. The accuracy of the classification of butterfly, flower seed and Brodatz datasets are 0.08, 0.3292 and 0.3343, respectively by TEM and 0.0053, 0.2417 and 0.3153, respectively by aTEM. The results of this study indicate that aTEM is a successful method for texture analysis.

## 1. Introduction

Image processing is an increasingly important area depends on recent developments in the quality of images, capabilities of computer and data storage systems. Also the usage of machine learning methods integrated with special analyzing machines has heightened the need for image processing techniques. Therefore the past thirty years have seen increasingly rapid advances in the field of image processing techniques and it has been used in various areas. One of the most important concepts of the image processing is texture analysis methods.

The texture was defined as a regular repetition of an element or a pattern [1] where a contiguous set of pixels with some tonal and/or regional property to provide information such as brightness, color, shape, size, etc [2]. Texture analysis has been used for recognizing or classifying objects, finding edges, shapes, etc. Texture methods was classified into three sub-groups: spectral methods transforms image into Fourier domain to detect global periodicity by identifying high energy, narrow peaks, statistical methods, analysis pixel-by-pixel relations and structural methods use properties of primitives and their placement. However, a major problem with this kind of applications is determining best suitable texture analysis method that increases meaningful of the image. A variety of methods are used to assess texture analyses. Each has its advantages and drawbacks. Therefore the texture analysis method must be chosen depending on accuracy, generalizing capacity, stability and computational cost. The popular texture analysis methods are gray-level co-occurrence matrix (GLCM), local binary patterns (LBP), texture energy measure (TEM) and histogram based features [3-8]. All texture methods have different characteristics and each one of them is more effective for some types of problems, such as in GLCM method the statistical features are changed depending on selected angle and distance parameters, while another statistical texture method, TEM can be used for detecting edges, levels, waves, spots and ripples at chosen vector length (VL) neighboring pixels in both horizontal and vertical direction [1].

TEM was defined as a similar process like the human visual process [3, 9] and it is based on filtering images by predefined special masks, which are obtained by the convolution of two 1-dimension (D) TEM vectors with either 3, 5 or 7 VL. TEM's masks were designed to detect spots, edges, level intensity (for 3 and 7 VL), ripple and wave (only for 5 VL) from both horizontal and vertical perspectives at the same time with a specified pixel neighbor.

Some statistical features of masked images, such as mean, standard deviation and entropy have extracted for analyzing images. A considerable amount of literature has been published about TEM because of being easily implemented and having better performance compare with other alternative approaches depend on providing several masked images that have different view to original image, therefore more useful features may be obtained from masked images [10]. Some of these studies are extracting features in ultrasonic liver images [11], detecting glandular tissue in breast X-rays [12], characterizing atherosclerotic carotid plaques [13], determining the composition of grain mixtures on industry [14] and also it was reported that TEM is one of the best methods for bone texture analysis [15]. Lemaitre and Rodojevic demonstrated that GLCM shows better results than TEM, but the computational cost of GLCM is higher than TEM [16]. On the other hand, Pietikainen et al. [17, 18] compared Laws, co-occurrence contrast, and edge per unit area operators on Brodatz and geological terrain types. They showed that TEM performed better than other operators. Also they point out that if the general forms of the masks were retained, performance did not deteriorate [18] which may be the major question of how to rise up the performance of TEM. Therefore, in 1983, Ade [19] proposed a method that is based on TEM in which it uses eigenfilters instead of TEM masks and the eigenfilters are occurred with all possible pairs of pixels in 3x3 masks, and later Unser improved this technique [20], on the other hand Vistnes reported that still the small scale masks may miss large scale textural structures [21].

One of the limitations of TEM is that its masks have not got the ability of suiting the image since it uses predefined masks. Another problem with this approach is that the masks are designed only for detecting spots, edges, level intensity (for 3 and 7 VL), ripple and wave (only for 5 VL) from only horizontal or vertical directions where the masks have not got the ability for detecting spots, edges, level intensity (for 3 and 7 VL), ripple and wave (only for 5 VL) of directions such as $45^0$.

The aim of this study was to evaluate and validate an improved version of TEM. Adaptive TEM method (aTEM) was defined and some images were used for showing the energy of images. In order to validate the performance of aTEM, the butterfly specie, Brodatz and flower seeds datasets were used. To assess aTEM also the TEM method was employed to these datasets and their accuracies were compared with each other and also with previous work results.

The rest of the paper is organized as follows. Section 2 gives an overview of TEM and describes aTEM, the improved TEM method. In this section, the characters of TEM and aTEM masks were discussed and masked images are also compared. Section 3 illustrates the experimental results of classification.   Section 4 draws conclusions.

## 2. Method

### 2.1. Texture Energy Measures

The TEM method is one of the most practical ways of texture analysis. TEM of an image is calculated by computing the energy of convolved images, with 2 dimension TEM masks, which are formed by convolving 1 D TEM vectors with each other. In order to determine levels, edges, ripples and spots in images, they are filtered with specific TEM masks. 3 types of 1 D TEM vectors are defined for each 3, 5 or 7 VL [3, 4 and 16] where the VL also refers neighboring size. The 1-D vectors are presented in Table 1.

*Table 1.* 1-D TEM Vectors

| Vector | VL 3 | VL 5 | VL 7 |
|---|---|---|---|
| Level | L3=[1 2 1] | L5 = [1 4 6 4 1] | L7=[1 6 15 20 15 6 1] |
| Edge | E3=[-1 0 1] | E5 = [-1 -2 0 3 1] | E7=[-1 -4 -5 0 5 4 1] |
| Spot | S3=[-1 2 -1] | S5 = [-1 0 2 0 -1] | S7=[-1 -2 1 4 1 -2 -1] |
| Wave | | W5 = [-1 2 0 -2 1] | |
| Ripple | | R5 = [1 -4 6 -4 1] | |

where level shows the average grey level and edge, spot, wave and ripple extracts edge, spots, waves and ripples in the image, respectively [4]. The TEM masks are determined by using 1-D TEM vectors and some samples of TEM masks with their calculations and descriptions are shown in table 2.

*Table 2.* TEM Mask Samples

| TEM Mask | Calculation | Description |
|---|---|---|
| $L_3E_3$ | $L_3^T E_3$ | Edge detection in horizontal direction and grey level intensity in vertical direction of 3 neighbouring pixels in both horizontal and vertical direction |
| $S_5S_5$ | $S_5^T S_5$ | Spot detection in both horizontal and vertical direction of 5 neighbouring pixels in both horizontal and vertical direction |
| $W_5R_5$ | $W_5^T R_5$ | Ripple detection in horizontal and wave detection in vertical direction of 5 neighbouring pixels in both horizontal and vertical direction |
| $S_7L_7$ | $S_7^T L_7$ | Grey level intensity in horizontal direction and spot detection in vertical direction of 7 neighbouring pixels in both horizontal and vertical direction |

There are 9 TEM masks that are defined by using combinations of 3 and 7 VL and 25 TEM masks that can be used with 5 VL. If the direction of the masks isn't important then the mean of the horizontal and vertical vectors of the same type of masks can be used for feature reduction, such as $S_7L_7TR = (S_7L_7 + L_7S_7)/2$ [16]. TEM masks with 7 VL are illustrated in Figure 1 [3].

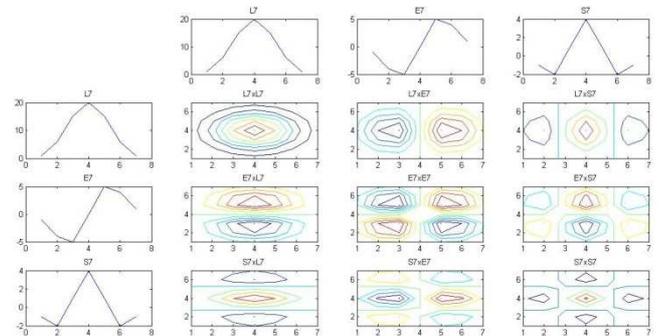

*Figure 1.* Masks with VL 7

Once the images were filtered with predefined TEM masks the energy of masked signals are calculated as follows: [4]

$$TEM(x,y) = \sum_{(u,v)\epsilon S_r(x,y)} |filtered\_image(u,v)| \quad (1)$$

where $(u,v)\epsilon S_r(x,y) \leftrightarrow \sqrt{(u-x)^2 + (v-y)^2} \leq r$ and $S_r(x,y)$ is the neighborhood around pixel $(x,y)$ with radius $r$. To classify images, the statistical features of the energized masked image are used. Generally the statistical features; mean, standard deviation, entropy, skewness and kurtosis can be used to describe the images [4], since it was reported that human texture perception is sensitive to first and second-order statistics and does not respond to higher than second-order [2].

### 2.2. Adaptive Texture Energy Measures (aTEM)

The main purpose of the study is to validate the applicability of the usage of adaptive vectors (AV) as a part of masks by convolving them with predefined 1 D TEM vectors. Later on the obtained masks can be used through different angles to the image for having different points of view. The new mask at $0^0$ detects level, edge, spot, ripple and wave in horizontal direction. The adaptive vector is expressed in eq. 2.

$$Adaptive vector = bias + \alpha * structural vector \quad (2)$$

where $bias$ shows the effect of contrast and $\alpha$ shows the sharpening of the mask. Any type of AVs can be defined depending on picking up any angle, contrast and sharpening parameters. The parameters must be determined due to the aim of analysis and the structure of the image. The structural vectors (SV) that used in this study are listed in Table 3.

*Table 3. aTEM Structural Vectors*

| Structural Vectors | VL 3 | VL 5 | VL 7 |
|---|---|---|---|
| Step | [1 1 1] | [1 1 1 1 1] | [1 1 1 1 1 1 1] |
| Ramp | [1 2 3] | [1 2 3 4 5] | [1 2 3 4 5 6 7] |
| Rectangular | [1 2 1] | [1 2 3 2 1] | [1 2 3 4 3 2 1] |
| Sinusoidal | | [1 2 1 2 1] | [1 2 1 2 1 2 1] |
| Sinusoidal2 | | | [1 2 3 2 1 2 3] |

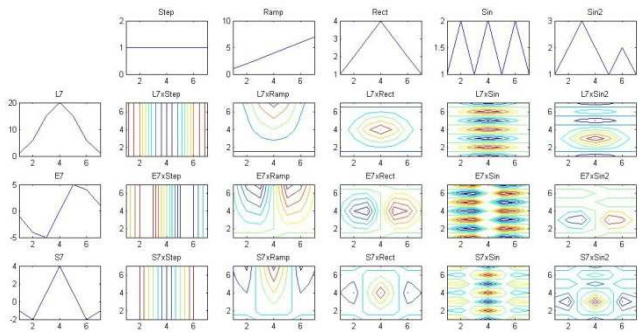

*Figure 2. aTEM Mask Samples with VL 7*

The SVs are determined due to their capability to define new masks into the desired structure by using 1-D TEM vectors. New aTEM masks were defined as; $A_n^T x T_n$, where $A$ is AV, $T$ is one dimensional TEM vector and $n$ is vector length, 3, 5 or 7. aTEM masks with 7 VL at $0^0$ with contrast (bias) is -3 and sharpening ($\alpha$) is 2 presented in figure 2.

Additionally, the aTEM masks can be rotate through the origin to define new filters by; $x' = x\cos\theta - y\sin\theta$ & $y' = x\sin\theta + y\cos\theta$, where $x'$ and $y'$ are new positions (coordinates) of x and y depend on rotating over the origin with θ angle. After "masking", the energy of masked signal is calculated by equation 1. For the purpose of classification, statistical features of the energy of the masked images can be used instead of images [4].

To determine the performance of aTEM; an image of Cancer Imaging Archive [22] was used. This sample was chosen because of the expected difficulty of obtaining edges. The sample, the TEM and aTEM masked image are shown in Figure 3 and Figure 4, respectively.

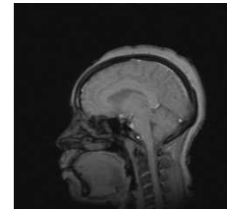

*Figure 3. An Image in Cancer Imaging Archive*

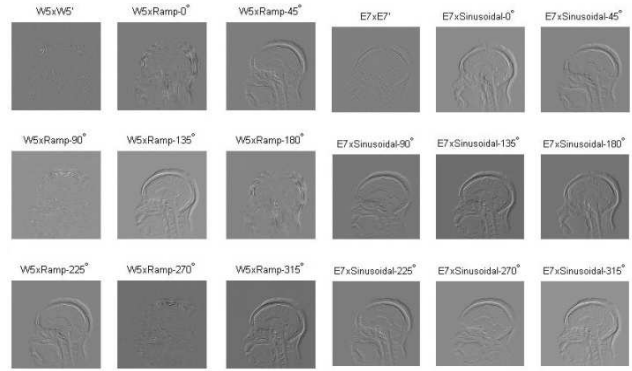

*Figure 4. TEM and aTEM Masked Samples of Cancer Image*

TEM and aTEM mapped sample, in figure 3, are showed in figure 4. Figure 4 compares the capacities of texture detections of TEM and aTEM for human vision and it is a good illustration of demonstrating the effect of the angle of the masks, since each different angle shows another perspective to the image. Generally the aTEM masked images are more understandable than TEM masked images and therefore aTEM masks are useful depend on the AV and the orientation angle of mask as seen in figure 4.

## 3. Materials

In this study butterfly identification [23], Brodatz [24] and follower seed datasets were used to classify for assessing the discrimination power of aTEM. Criteria for selecting these three subjects were as follows: the butterfly data-set [23] is a complex dataset for classification. The Brodatz dataset [24] is a benchmark dataset for image processing methods and the last dataset, flower seed dataset, can be an interesting example for texture analysis.

### 3.1. Butterfly Dataset

The butterflies, belonging to family Papilinidae, was collected from Mount Erek, Van between May 2002 and August 2003 in the attitudes of 1800-3200 meters, were used for identification of texture features of the species. The dataset consisted of 10 images for each of 19 butterfly species that was shot with a Nikon Professional camera [23]. The butterfly images used in the study are shown in Figure 5.

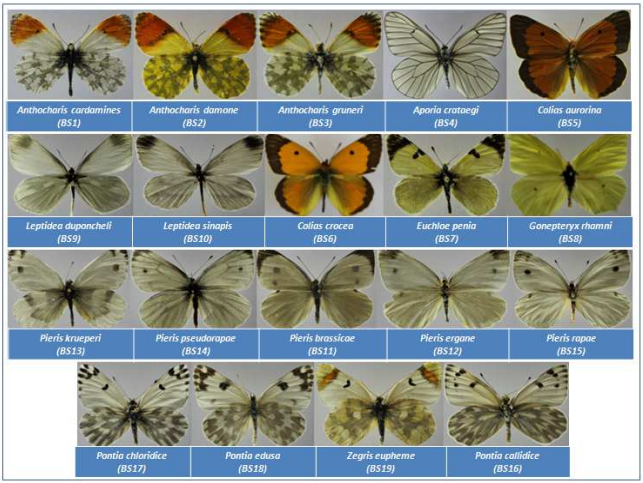

*Figure 5.* The selected samples from nineteen butterflies species

### 3.2. Flower Seed Dataset

This dataset is consisted of 240 images of 6 different flower seed specie and the image was recorded on an Olympus SZ61 stereo microscope integrated Olympus DP20 camera. Selected samples of the flower seed images used in this study are shown in Figure 6.

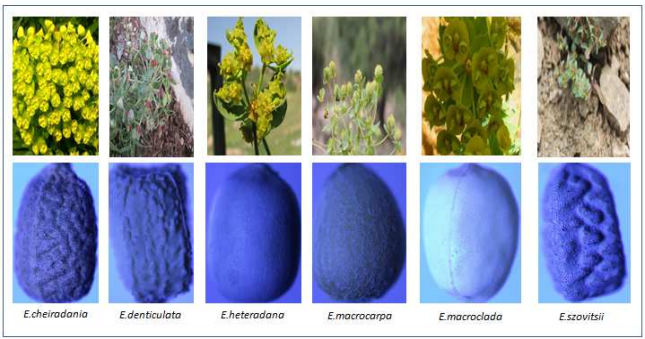

*Figure 6.* The selected samples from six flower seed

### 3.2. Brodatz Dataset

Another example used for assessing aTEM is the Brodatz Dataset [24] that is shown in Figure 7. The Brodatz dataset, which the images suffer a very high signal to noise ratio and formed under different imaging conditions have been utilized by many researchers [25].

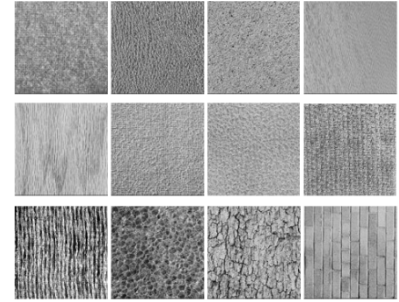

*Figure 7.* The selected samples from Brodatz Dataset

## 4. Results and Discussion

This study set out with the aim of assessing the performance of the proposed method aTEM which is an improved version of TEM in texture analysis. Therefore the experiments in present study were designed to determine a comparison with aTEM and TEM and also with the previous studies.

The mean, standard deviation and entropy of the masked images was used as features of the images. In order to assess the accuracies of classification with extracted features by TEM and aTEM, k Nearest Neighbor (kNN), a popular and easy machine learning method, was used [26]. kNN depends on a simple idea, which is the class of an unclassified data is the same of its nearest neighbors class. The main problem of using kNN is choosing the best suitable number of neighbor and distance calculation method such as Euclidian, Manhattan, Supreme and etc… In this study the number of nearest neighbor and distance calculation method are picked out as 1 and Euclidian distance calculation method, respectively. The error is calculated by:

$$Error = \#\{Wrong\ Classified\ Data\}/\#\{Dataset\}$$

Furthermore these experiments were done with 10-folds crossvalidation to minimize the error of the data distribution of dataset. The average errors of aTEM and TEM were compared in order to validate the applicability of aTEM as a texture analysis method is shown in table 4.

*Table 4.* Classification Errors

|  |  | Butterfly | Flower Seed | Brodatz |
|---|---|---|---|---|
| TEM | VL 3 | 0.08 | 0.3292 | 0.3343 |
|  | VL5 | 0.043 | 0.4292 | 0.3594 |
|  | VL 7 | 0.08 | 0.3875 | 0.4254 |
| aTEM | Min | 0.0053 | 0.2417 | 0.3153 |
|  | Mean | 0.0141 | 0.2802 | 0.3558 |
|  | Max | 0.0263 | 0.3917 | 0.4605 |

Table 4 represents the results obtained from the classification of datasets. As can be seen from the table (above), the results of aTEM is reported significantly higher accuracy than the results obtained by employing TEM. The minimum errors obtained by the TEM method for butterfly, flower seed and Brodatz datasets are 0.043, 0.3292 and 0.3343, respectively and by aTEM are 0.0053, 0.2458 and 0.3153, respectively. As seen from the results, the features extracted by aTEM have higher accuracy than the features extracted by TEM.

The lowest and highest classification error is obtained when employing VL 7 aTEM masks, and VL 5 aTEM masks that formed with the step SV in butterfly dataset. In order to show how error regulates depends on contrast and sharpening parameters, a series of classification was performed and the error curve is illustrated in figure 8 while the 5 VL aTEM masks are formed by the rectangular SV and window size was 10. As seen in Fig. 8 there is not any clear correlation with contrast and sharpening parameters between errors.

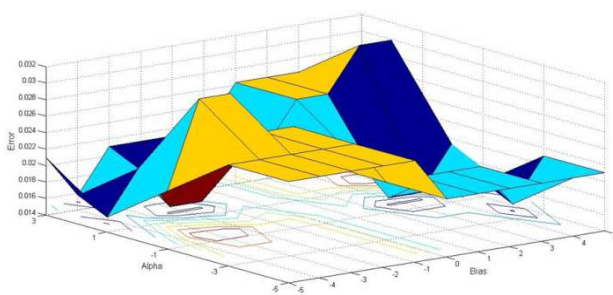

*Figure 8. Error Curve of a TEM Masked Butterfly Image Classification*

The obtained minimum error in Kaya et al. [27] study is 0.0125 (1,25%) with GLCM (feature extraction) and extreme learning machine (classification method). In their study the both GLCM and LBP methods were used for feature extraction and the classifications were carried out with support vector machine, Bayesian network, logistic regression, artificial neural network, decision rules algorithm, decision tree algorithm (J48) and extreme learning machine models. This study also produced results which corroborate the findings of Pietikainen et al. [17, 18] in this field that TEM shows higher accuracy than GLCM and LBP.

In flower seed dataset; the minimum classification error is obtained while employing with VL 5 aTEM that is formed with rectangular SV and the sharpening parameter is -1 and the highest error is obtained while using VL 3 aTEM that is formed with ramp SV and sharpening parameter is 1.

The results of Brodatz dataset is for lowest and highest classification error obtained while employing VL 5, VL 3 aTEM masks, respectively. The error of the results obtained by aTEM is lower than TEM as seen in Table 4. In previous studies 0.2085 and 0.2879 errors are obtained with one and two level wavelet decompositions of images, respectively [28]. The errors obtained by aTEM are higher than the results in [28], which may be because of the used classification method which is not optimized and the high signal to noise ratios of images.

As a summary Table 4 shows that there is a significant accuracy difference between the classification results of aTEM and TEM. The minimum error results were showing that aTEM is a successful feature extractor. There are several possible explanations for these results. A possible explanation for this might be that the number of masks employed in aTEM is much more than the masks that can be used in TEM, where this finding is in agreement with the findings of Pietikainen et al [17, 18]. Another possible explanation for this is that the masks in aTEM can be adapted to image by determining the best orientation angle, contrast and sharpening of the mask

Although aTEM showed a great success of classification, in this method there is a major problem with determining the best angle, contrast and sharpening parameters which can be obtained by trials or an expert view as the same in determining optimal artificial neural network parameters process. In this study it was not possible to investigate the significant relationships of angle, contrast and sharpening parameters with image characteristics. It is somewhat surprising that no effect of these parameters was noted in some experiments. In general, therefore, it seems that the time cost of aTEM is higher than TEM in determining best parameters, but once the optimal parameters are determined the aTEM is as fast as TEM.

## 5. Conclusion

Prior studies have noted the importance of texture analyses and the TEM method. In this study a new method, Adaptive Texture Energy Measure Method (aTEM) which is based on Law's TEM method was offered. aTEM uses AVs for forming new masks by convolving these AVs with 1 D TEM vectors. The detection capability of the aTEM masks used for cancer image, were found satisfactory. Butterfly species identification, follower seed classification and Brodatz Dataset classification examples were shown that aTEM is a successful method for classification. In the current study, comparing the classification errors of features extracted by aTEM with TEM showed that the aTEM is more successful than TEM. There are a number of important improvements of aTEM from TEM. The aTEM masks are adaptive due to contrast and sharpening and also the orientation angle of the mask. The present results are significant in at least major one respect. Employing image adapted masks to images shows higher accuracy than predefined masks which is important for developing expert texture analysis methods. This research will serve as a base for future studies and the special masks may be determined for any type of image dataset.